\documentclass[lettersize,journal]{IEEEtran}
\usepackage{acronym}
\newacro{CAV}{connected and automated vehicle}
\newacro{UAV}{unmanned aerial vehicle}
\newacro{SDR}{semidefinite relaxation}
\newacro{SDP}{semidefinite programming}
\newacro{NLP}{nonlinear program}
\newacro{OCP}{optimal control problem}
\newacro{MPC}{model predictive control}
\newacro{NMPC}{nonlinear model predictive control}
\newacro{QCQP}{quadratically constrained quadratic program}

\usepackage{cite}
\usepackage{amsmath,amssymb,amsfonts}

\usepackage{graphicx}
\usepackage{textcomp}
\usepackage{tabu}
\usepackage{booktabs}
\usepackage{multirow}
\usepackage{amsmath}

\newacro{LTI}{linear time-invariant}

\usepackage{epstopdf}
\usepackage{pbox}
\usepackage{algorithm}
\usepackage{CJK}
\usepackage{mathtools}%

\newcounter{step}
\renewcommand{\thestep}{S\arabic{step}}

\newcommand{\skipthis}[1]{ }

 \usepackage{color}

\usepackage[short]{optidef}
\usepackage{amsmath}
\usepackage[compatible]{algpseudocode}
\usepackage{multirow}
\usepackage[normalem]{ulem}
\useunder{\uline}{\ul}{}
\usepackage{booktabs} 
\usepackage{siunitx}  
\usepackage[table]{xcolor} 
\usepackage{flushend}
\usepackage{comment}
\usepackage{tcolorbox}
\usepackage{url}
\newtcolorbox{algbox}[2][]{%
	colframe=gray!15, 
	colback=gray!05, 
	coltitle=black, 
	fonttitle=\bfseries,
	colupper=black, 
	title=#2, 
	arc=1mm,
	boxrule=0.0mm, 
	left=0pt,
	right=0pt,
	top=-5pt,
	bottom=0pt,
	#1 
}

\usepackage[breaklinks,colorlinks,linkcolor=red,citecolor=blue]{hyperref}

\begin{document}

\title{A Fast Semidefinite Convex Relaxation for Optimal Control Problems With Spatio-Temporal Constraints}

\author{Shiying Dong, Zhipeng Shen, Rudolf Reiter, Hailong Huang, Bingzhao Gao, \\ Hong Chen, \emph{IEEE Fellow}, and Wen-Hua Chen, \emph{IEEE Fellow}

\thanks{This work was supported by the Research Centre for Low Altitude Economy (RCLAE), the Hong Kong Polytechnic University. (Corresponding authors: Bingzhao Gao and Wen-Hua Chen)}
\thanks{S. Dong, Z. Shen, H. Huang, and W. H. Chen are with the Department of Aeronautical and Aviation Engineering, the Hong Kong Polytechnic University, Hong Kong, China. (\{shiying.dong, hailong.huang, wenhua.chen\}@polyu.edu.hk, zhipeng.shen@connect.polyu.hk)}
\thanks{R. Reiter is with the Robotics and Perception Group, University of Zurich, Switzerland. (rreiter@ifi.uzh.ch)}
\thanks{B. Gao is with the School of Automotive Studies, Tongji University, Shanghai, China. (gaobz@tongji.edu.cn)}
\thanks{H. Chen is with the College of Electronic \& Information Engineering, Tongji University, Shanghai, China. (chenhong2019@tongji.edu.cn)}
\thanks{}

}


\maketitle

\begin{abstract}
Solving \acp{OCP} of autonomous agents operating under spatial and temporal constraints fast and accurately is essential in applications ranging from eco-driving of autonomous vehicles to quadrotor navigation.
However, the nonlinear programs approximating the \acp{OCP} are inherently nonconvex due to the coupling between the dynamics and the event timing, and therefore, they are challenging to solve. 
Most approaches address this challenge by predefining waypoint times or just using nonconvex trajectory optimization, which simplifies the problem but often yields suboptimal solutions. 
To significantly improve the numerical properties, we propose a formulation with a time-scaling direct multiple shooting scheme that partitions the prediction horizon into segments aligned with characteristic time constraints. 
Moreover, we develop a fast semidefinite-programming-based convex relaxation that exploits the sparsity pattern of the lifted formulation. 
Comprehensive simulation studies demonstrate the solution optimality and computational efficiency. 
Furthermore, real-world experiments on a quadrotor waypoint flight task with constrained open time windows validate the practical applicability of the approach in complex environments.

\end{abstract}


\maketitle

\acresetall 

\section{Introduction}

In recent years, it has become increasingly evident that time- and/or energy-optimal control is essential for the broader deployment of autonomous agents such as \acp{CAV}~\cite{reiter2024equivariant,dong2024real,allamaa2024learning} and \acp{UAV}~\cite{krinner2024mpcc++,wang2025unlocking}. 
Among the wide range of challenges in this domain, a notable class of problems involves autonomous agents navigating predefined routes that include a series of gates or waypoints. 
These problems are critical to applications such as infrastructure inspection, logistics and delivery, search and rescue missions, drone racing, and eco-driving of \acp{CAV} where traffic light signal information can be exploited to minimize stop-and-go maneuvers \cite{kaufmann2023champion,foehn2021time,vahidi2018energy,han2019fundamentals}.  

However, optimizing the motion of an autonomous agent to cross multiple gates is nontrivial. 
The timely crossing of these gates, which correspond to signalized intersections in ground vehicles or aerial waypoints for \acp{UAV}, naturally imposes multiple spatio-temporal constraints. 
These constraints include: (i) spatial equality constraints that ensure the trajectory passes through the exact physical location of each gate, and (ii) temporal inequality constraints representing the open time windows (e.g., green light phases, safe passing intervals). 
Furthermore, the task must be completed in a manner that minimizes time and/or energy. 
To fully exploit the potential for minimizing such costs, a solution must simultaneously optimize both the agent’s trajectory and the crossing times at each gate.
However, when spatio-temporal constraints are imposed, the resulting optimization problem becomes nonconvex and challenging to solve, even for locally optimal solutions.

A variety of approaches have been proposed to tackle this problem. 
In the eco-driving context for \acp{CAV}  approaching multiple signalized intersections, one popular strategy is to predefine the gate crossing times and then optimize the vehicle’s speed profile accordingly \cite{ard2025energy}. 
This effectively reduces the problem to one involving only spatial constraints, but such simplification often precludes lower-cost solutions that could arise from flexible timing, resulting in reduced efficiency or infeasible plans. 
To address this limitation, some works have attempted to co-optimize crossing times and speed trajectories. 
For instance, De Nunzio \emph{et al.} \cite{de2016eco} discretized the feasible passing times at each intersection and formulated the problem as a shortest-path search, while Dong \emph{et al.} \cite{dong2022predictive} developed a hierarchical framework for energy-efficient driving that incorporates traffic light information into speed profile optimization. 
However, these methods are largely restricted to two-dimensional (2D) road vehicle maneuvers, limiting their applicability to complex 3D scenarios.

In the \ac{UAV} domain, quadrotor waypoint flight or drone racing with gates has received significant research attention \cite{hanover2024autonomous}. 
Polynomial planning is a common strategy that leverages differential flatness~\cite{richter2016polynomial}. 
Building on these foundations, the authors in \cite{qin2023time} adapted the framework for drone racing, addressing collision avoidance and gate traversal in dynamic environments.
The paper \cite{zhou2023efficient} introduces a novel method for computing time-optimal trajectories, in which the nodes subject to waypoint constraints are fixed, while distinct sampling intervals are assigned to the trajectory segments between successive waypoints.
The paper \cite{shen2024sequential} proposes a sequential convex programming-based algorithm to reduce the computational burden and improve the solution quality of quadrotor waypoint flight.
Although the past decade has seen substantial progress in waypoint-based quadrotor trajectory planning, the more challenging problem of jointly enforcing spatial waypoint constraints and temporal access windows at each waypoint has remained largely unexplored.  
Some numerical optimization-based approaches \cite{dong2023cooperative,meng2020trajectory} attempt to jointly minimize time and energy consumption under spatio-temporal constraints. 
Yet, these methods typically reformulate the original problem as a nonconvex \ac{NLP} and rely on local solvers. 
As a result, they are sensitive to initial guesses and prone to being stuck in local optima.

Recent advancements in optimal control and convex relaxation have opened up new possibilities for solving such problems more effectively. 
Loxton \emph{et al.} \cite{loxton2008optimal} formalized a general class of \acp{OCP} whose objectives and constraints depend on multiple discrete time points. 
A computational method using time-scaling transformation \cite{teo2002numerical} was subsequently proposed to address this class of problems. 
However, when trajectory and time-scaling variables are optimized jointly, the resulting problem is inherently nonconvex and challenging to solve efficiently. 
Meanwhile, advances in \ac{SDR} have demonstrated strong potential for transforming certain nonconvex optimal control and planning problems into tractable convex programs \cite{marcucci2025biconvex,graesdal2024convexrelaxations}. 
For example, \cite{graesdal2024convexrelaxations} used \ac{SDR} to relax quasi-static nonconvex dynamics in a fixed-time planar pushing task, and Yang \emph{et al.} \cite{yang2025new} proposed a structure-exploiting SDR for time-scaled optimal control.  
While SDR has been successfully applied to motion planning problems, extending it to optimization under spatio-temporal constraints has not been studied.

Motivated by these discussions, we address for the first time the global \ac{OCP} with spatio-temporal constraints. 
To generalize and systematically address such problems, we formulate them as a class of \acp{OCP} with characteristic time constraints. 
To enable efficient numerical computation, we develop a tailored numerical approach that integrates time-scaling with a direct multiple-shooting scheme, yielding structured yet inherently nonconvex \acp{NLP} that seamlessly incorporate both trajectory dynamics and spatio-temporal constraints within a unified formulation.
We then propose a problem-specific, fast \ac{SDP}-based convex relaxation that exploits the sparsity and structure of the time-scaled formulation to tightly approximate the nonconvex components while remaining computationally tractable.
The main contributions of this paper are as follows:
\begin{enumerate}
    \item This paper develops a tailored problem formulation and numerical approach that enables the simultaneous optimization of trajectory and gate-crossing times, providing practical solutions for complex spatio-temporal motion planning tasks.
    \item We introduce a sparse-structure-aware, problem-specific fast SDR method for the formulated nonconvex \ac{OCP} with spatio-temporal constraints, achieving tight global approximations while reducing computational cost.
\end{enumerate}
We demonstrate the proposed method’s applicability in both \ac{CAV} and UAV domains, including real-world experiments on quadrotor waypoint flight under time-window constraints.

\section{Preliminaries on Semidefinite Programming Relaxation}\label{} 

Quadratically Constrained Quadratic Programs (QCQPs) \cite{park2017general} are a class of optimization problems in which both the objective function and the constraints are quadratic in the decision variables. 
Without loss of generality, a generic QCQP \cite{luo2010semidefinite} is formulated as
\begin{equation}
\begin{aligned}
\min_{x \in \mathbb{R}^n} \quad & x^\top Q x \\
\text{s.t.} \quad & x^\top P_i x \geq b_i, \quad i = 1, \ldots, m,
\end{aligned}
\label{eq:qcqp}
\end{equation}
where $x \in \mathbb{R}^n$ is optimization variable, and \( Q, P_1, \ldots, P_m \in \mathbb{S}^n \) are real symmetric matrices.  
The matrix \( Q,  P_i,i = 1, \ldots, m \) may be indefinite, rendering the problem nonconvex and, in general, NP-hard.

Using the trace identity \( x^\top Q x = \mathrm{trace}(x^\top Q x) = \mathrm{trace}(Q x x^\top) \), we can express the quadratic term as
\begin{equation}
x^\top Q x = \mathrm{trace}(Q x x^\top).
\label{eq:trace}
\end{equation}
\begin{equation}
x^\top P_i x = \mathrm{trace}(P_i x x^\top), \quad i = 1, \ldots, m.
\label{eq:trace}
\end{equation}

A popular approach to address the nonconvexity of QCQPs is via SDR.
 By introducing a new lifted positive semidefinite (PSD) matrix variable \( X = x x^\top \), the quadratic terms can be linearized using the trace identity. This leads to the reformulated problem:
\begin{equation}
\begin{aligned}
\min_{X \in \mathbb{S}^n} \quad & \mathrm{trace}(Q X) \\
\text{s.t.} \quad & \mathrm{trace}(P_i X) \geq b_i, \quad i = 1, \ldots, m, \\
& X \succeq 0, \\ 
& \mathrm{rank}(X) = 1.
\end{aligned}
\label{eq:qcqp-lifted}
\end{equation}

The formulation~\eqref{eq:qcqp-lifted} is still nonconvex due to the rank constraint. 
By relaxing the rank-one constraint, we obtain a standard SDP relaxation of the original QCQP:
\begin{equation}
\begin{aligned}
\min_{X \in \mathbb{S}^n} \quad & \mathrm{trace}(Q X) \\
\text{s.t.} \quad & \mathrm{trace}(P_i X) \geq b_i, \quad i = 1, \ldots, m, \\
& X \succeq 0.
\end{aligned}
\label{eq:sdp-relax}
\end{equation}

This relaxed problem~\eqref{eq:sdp-relax} is a convex optimization problem and can be solved efficiently using interior-point methods. 
The solution to the relaxed problem serves as a lower bound (for minimization) to the original QCQP.
Although the relaxation may not always yield a rank-one solution, in many structured cases arising in control and estimation problems, the relaxation is tight or sufficiently accurate for practical use.

\section{General Problem Formulation}\label{} 

We consider a trajectory optimization problem for an autonomous agent operating under spatio-temporal constraints, which commonly arise in applications such as eco-driving of \acp{CAV}~\cite{ard2025energy,dong2023cooperative} and quadrotor waypoint flight with time-window constraints~\cite{foehn2021time,zhou2023efficient}. 
The agent is required to sequentially pass through \( n \) waypoints with 3D coordinates \( \mathbf{p}_i \in \mathbb{R}^3 \), \( i = 1, \dots, n \). 
Each waypoint \( i \) is associated with an admissible open-time interval \( T_i^\circ \subset \mathbb{R}^+ \), and the agent may only cross the waypoint at \( t_i \in T_i^\circ \). 

The agent dynamics follow a second-order system:
\begin{align}
\dot{\mathbf{p}}(t) &= \mathbf{v}(t), &
\dot{\mathbf{v}}(t) &= \mathbf{a}(t),
\end{align}
where \( \mathbf{p}(t) = [p_x(t), p_y(t), p_z(t)]^\top \) is position, 
\( \mathbf{v}(t) = [v_x(t), v_y(t), v_z(t)]^\top \) is velocity, 
and \( \mathbf{a}(t) = [a_x(t), a_y(t), a_z(t)]^\top \) is the control input (acceleration). 
Velocity and acceleration are bounded component-wise:
\begin{equation}
 v_{\min}  \leq \mathbf{v}(t) \leq  v_{\max} , \quad 
 a_{\min}  \leq \mathbf{a}(t) \leq a_{\max}.
\end{equation}
The objective is to minimize both total time and control effort:
\begin{equation}
J = \rho_t (t_f - t_0) + \rho_u \int_{t_0}^{t_f} \|\mathbf{a}(t)\|_2^2 \, \mathrm{d}t,
\end{equation}
subject to spatial and temporal constraints:
\begin{equation}
\mathbf{p}(t_i) = \mathbf{p}_i, \quad t_i \in T_i^\circ, \quad t_{i+1} > t_i.
\end{equation}

This formulation defines a trajectory optimization with spatio-temporal constraints problem, which can be viewed as a specific case of a more general \ac{OCP} with characteristic time constraints.

To generalize, consider a linear time-invariant (LTI) system:
\begin{equation}
\dot{x}(t) = A x(t) + B u(t), \quad x(0) = \bar{x}_0, \; x(t_f) = \bar{x}_f,
\end{equation}
subject to path bounds and characteristic time constraints:
\begin{small}
\begin{equation}
\psi_l(x(\tau_l)) := x(\tau_l) - \bar{x}_l = 0, \quad 
\tau_{l,\min} \leq \tau_l \leq \tau_{l,\max}, \; l = 1, \dots, M.
\end{equation}
\end{small}

The cost function balances total trip time and energy use
leading to the following compact form:
\begin{mini!}[1]
{\substack{x(\cdot), u(\cdot), t_f, \{\tau_l\}}}
{J = w_t t_f + \int_0^{t_f} \left( u^\top Q u + x^\top R x \right) \mathrm{d}t}
{\label{eq:ocp-ctc}}{}
\addConstraint{ x(0) - \bar{x}_0 = 0, x(t_f) - \bar{x}_f} {= 0}
\addConstraint{ \dot{x}(t) - \left( A x(t) + B u(t) \right)} {= 0, & \quad t \in [0, t_f]} \label{}
\addConstraint{x_{\max} \geq  x(t) } {\geq  x_{\min}, & \quad t \in [0, t_f]} \label{}
\addConstraint{u_{\max} \geq  u(t) } {\geq u_{\min}, & \quad t \in [0, t_f]} \label{}
\addConstraint{\tau_{l,\max}  \geq  \tau_l} {\geq \tau_{l,\min}, & \quad t \in [0, t_f]} \label{}
\addConstraint{\psi_l(x(\tau_l))} {= 0, & \quad l = 1, \dots, M,} \label{}
\end{mini!}

\section{Convex Optimization for Nonconvex OCPs}\label{}

\subsection{Direct Multiple Shooting and Time-scaling Formulation}\label{} 


\begin{figure}[t]
	\centering
		\includegraphics[width=7.5cm]{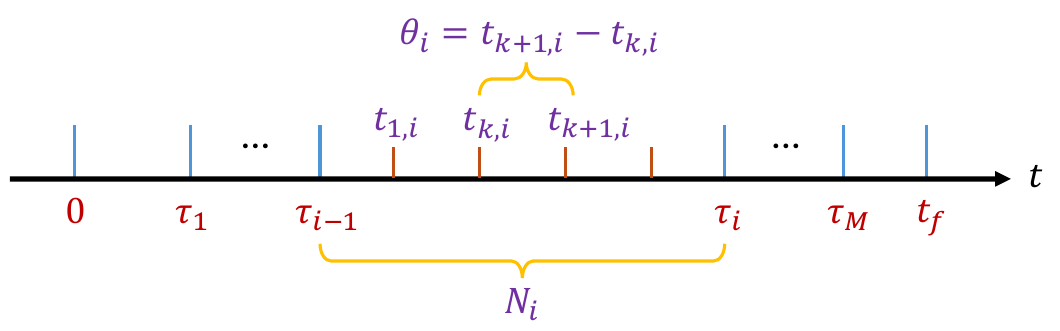}
    	  \caption{Time-scaling illustration. } \label{timescaling}
\end{figure}

To enable efficient numerical solutions to \ac{OCP} with characteristic time constraints, we develop a tailored approach that combines time-scaling with a direct multiple shooting scheme. 
As illustrated in Fig. \ref{timescaling}, the \( M \) characteristic times \( \tau_1, \dots, \tau_M \) divide the overall time horizon \( [0, t_f] \) into \( M+1 \) sequential segments. 
To discretize the problem, we apply direct multiple shooting \cite{bock1984multiple} by assigning a fixed number \( N_i \) of shooting intervals to each segment \( i = 1, \dots, M+1 \), resulting in a total of $N = \sum_{i=1}^{M+1} N_i$ shooting intervals across the full horizon.
Unlike traditional formulations \cite{bock1984multiple}, \cite{dong2023cooperative} that assign a separate time step to each shooting interval, we introduce a more compact time-scaling representation: each segment \( i \) is associated with a single scalar variable \( \theta_i > 0 \). 
Consequently, each segment is uniformly subdivided into \( N_i \) intervals of equal length:
\begin{equation}
\theta_{i} := t_{k+1,i} - t_{k,i},  k = 0, \dots, N_i - 1, \quad i = 1, \dots, M+1,
\end{equation}
which are treated as an independent optimization variable. 
These time-step variables are constrained to be non-negative, i.e., $\theta_{i} \geq 0, \quad i = 1, \dots, M+1$.
The total final time becomes:
\begin{equation}
t_f = \sum_{i=1}^{M+1} N_i\theta_i.
\end{equation}
Similarly, each characteristic time \( \tau_i \), which marks the end of segment \( i \), is given by the accumulated duration up to that segment:
\begin{equation}
\tau_i = \sum_{l=1}^{i} N_l \theta_l, \quad i = 1, \dots, M.
\end{equation}

This segment-wise time-scaling approach introduces only \( M+1 \) scalar time variables while still allowing flexible timing across the entire horizon.

To discretize the dynamics, we assume the control input is piecewise constant over each interval $u(t) = u_{k,i}, \quad \forall t \in [t_{k,i}, t_{k+1,i})$, and integrate the dynamics numerically over each interval to obtain:
\begin{equation} \label{direct shooting dynamics}
x_{k+1,i} = F_{k,i}(x_{k,i}, u_{k,i}, \theta_i),
\end{equation}
where \( F_k \) represents the result of applying a numerical integrator (e.g., Runge-Kutta or Euler) to the system dynamics.
In this work, we consider the forward Euler scheme for simplicity:
\begin{equation}
F_{k,i}(x_{k,i}, u_{k,i}, \theta_i) = x_{k,i} + \theta_i (A x_{k,i} + B u_{k,i}).
\end{equation}

The continuous-time cost functional is approximated by a sum over the intervals:
\begin{equation}
\sum_{i=0}^{M+1}w_t N_i\theta_i + \sum_{i=0}^{M+1}\sum_{k=0}^{N_i-1} L_{k,i}(x_{k,i}, u_{k,i}, \theta_i),
\end{equation}
with \( L_{k,i} \) being a numerical approximation of the integral cost over interval \( [t_{k,i}, t_{k+1,i}] \). 
In the case of using the forward Euler method, \( L_{k,i}  \) can be approximated as:
\begin{equation}
L_{k,i}(x_{k,i}, u_{k,i}, \theta_i) = \theta_i \left( x_{k,i}^\top Q_x x_{k,i} + u_{k,i}^\top R_u u_{k,i} \right),
\end{equation}
where \( Q_x \succeq 0 \) and \( R_u \succeq 0 \) are the control and state cost weighting matrices, respectively. 

Moreover, to ensure temporal consistency of the trajectory across segment boundaries, the terminal state of segment \( i \) must match the initial state of segment \( i+1 \). This inter-segment continuity condition is enforced by the following constraints:
\begin{equation} \label{segment continuity condition}
x_{N_i, i} = x_{0, i+1}, \quad  i = 1, \dots, M.
\end{equation}
Here, \( x_{N_i, i} \) denotes the terminal state obtained after integrating over the last shooting interval in segment \( i \), while \( x_{0, i+1} \) is the initial state of the first interval in segment \( i+1 \).
These equality constraints guarantee that the entire trajectory across the \( M+1 \) segments forms a continuous and dynamically feasible path. 

Through the use of direct multiple shooting and a time-scaling transformation, the continuous-time \ac{OCP} \eqref{eq:ocp-ctc} with characteristic time constraints is formulated into the following finite-dimensional \ac{NLP} problem:

\begin{small}
\begin{mini!}[1]
{\mathbf{x}, \mathbf{u},\mathbf{\theta}}{\sum_{i=0}^{M+1}w_t N_i\theta_i + \sum_{i=0}^{M+1}\sum_{k=0}^{N_i-1} L_{k,i}(x_{k,i}, u_{k,i}, \theta_i),
}
{\label{original QCQP}}{}
\addConstraint{x_{0,1} = \bar{x}_0, x_{N_{M+1},M+1}} {= \bar{x}_f}
\addConstraint{x_{k+1,i} - F_{k,i}(x_{k,i}, u_{k,i}, \theta_i)} {= 0, & i = 1, \dots, M+1} \label{}
\addConstraint{x_{\max} \geq  x_{k,i}} {\geq x_{\min}, & i = 1, \dots, M+1} \label{}
\addConstraint{u_{\max} \geq u_{k,i}} {\geq u_{\min},  & i = 1, \dots, M+1} \label{}
\addConstraint{\theta_i} {\geq 0, & i = 1, \dots, M+1} \label{}
\addConstraint{x_{N_i, i} - x_{0, i+1}} {= 0, & i = 1, \dots, M} \label{}
\addConstraint{\psi_l(x_{N_i,i})} {= 0, & i = 1, \dots, M} \label{}
\addConstraint{{\tau}_{i,\max}\geq \sum_{l=1}^{i} N_l \theta_l} {\geq {\tau}_{i,\min}, & i = 1, \dots, M} \label{}
\addConstraint{} {& k = 0, \dots, N_i-1, \notag} 
\end{mini!}
\end{small}
\noindent  where $\mathbf{x} = \left(x_{0,1}, \ldots, x_{N_1,1}, x_{0,2},  \ldots, x_{N_{M+1},M+1} \right)$, $\mathbf{u} = \left(u_{0,1}, \ldots, u_{N_1-1,1}, u_{0,2},  \ldots, u_{N_{M+1}-1,M+1} \right)$, $\mathbf{\theta} = \left(\theta_1, \ldots, \theta_{M+1} \right)$.

\subsection{Semidefinite Programming-based Convex Relaxation}\label{} 

For the sake of notational simplicity, in the remainder of this section we omit the index \( i \) from the state and control variables, and denote them simply as \( x_k \) and \( u_k \) for each shooting node \( k \). 
The time-scaling variable \( \theta_i \), which is defined per segment, retains its index to maintain clarity regarding its association with segment \( i \).
The resulting  \ac{NLP} problem \eqref{original QCQP} remains inherently nonconvex. 
This nonconvexity arises  from the bilinear coupling between the variables \( \theta_k \) and the variables \( x_k, u_k \) in both the dynamic equations and the objective function. 

The cost function contains terms like \( \theta_i ( x_{k}^\top Q_x x_{k} )\) and $\theta_i (  u_{k}^\top R_u u_{k})$, which go beyond the classical quadratic structure handled by conventional SDR.
Nonetheless, the problem exhibits a valuable structure that can be exploited to design a tractable convex relaxation. 
Due to the use of direct multiple shooting, the coupling between variables is sparse: each stage cost and constraint depends only on a small subset of variables, namely \( w_k= (x_k, x_{k+1}, u_k) \). 
This localized interaction is a crucial property for reducing computational burden and obtaining scalable semidefinite optimization.
To address the nonconvexity caused by the bilinear and polynomial terms such as \( \theta_i x_k^\top x_k \), \( \theta_i u_k^\top u_k \), \( \theta_i x_k \), and \( \theta_i u_k \), we introduce a lifting strategy that embeds the bilinear structure into a higher-dimensional space. Define the following lifted variable:

\begin{small}
\begin{equation}
    w_k= \begin{bmatrix} x_k \\ x_{k+1} \\ u_k  \end{bmatrix}, \quad y_k = \begin{bmatrix}1 \\ \theta_i \\ w_k  \\ \theta_kw_k \end{bmatrix},
\end{equation}
\begin{equation}
Y_{k,i} = y_k y_k^\top =
\begin{bmatrix}
1 & \theta_i & w_k^\top & \theta_k w_k^\top \\
\theta_i & \theta_i^2 & \theta_i w_k^\top & \theta_i^2 w_k^\top \\
w_k & \theta_i w_k & w_k w_k^\top & \theta_i w_k w_k^\top \\
\theta_i w_k & \theta_i^2 w_k & \theta_i w_k w_k^\top & \theta_i^2 w_k w_k^\top
\end{bmatrix}
\end{equation}
\end{small}
This lifted PSD matrix \( Y_{k,i} \in \mathbb{S}^{r}_+, r= 2+4n_x+2n_u \) captures the monomials \( x_k^\top x_k \), \( u_k^\top u_k \), and their cross-terms with \( \theta_i \). However, it also introduces higher-order terms such as \( \theta_i^2 \) and \( \theta_i^2 x_k x_k^\top \), which are not present in the original formulation and therefore may lead to an overly loose relaxation.

To overcome this challenge, we propose a structure-preserving semidefinite lifting scheme tailored to the specific bilinear form of the \ac{OCP} objective. 
Noting that \( \theta_i > 0 \) by construction, we adopt a transformation inspired by \cite{yang2025new}, which introduces a scaled lifted PSD matrix:

\begin{small}
\begin{equation}
\tilde{Y}_{k,i} = \frac{1}{\theta_i} y_k y_k^\top = \begin{bmatrix}
\frac{1}{\theta_i} & 1 & \frac{1}{\theta_i} w_k^\top & w_k^\top \\
1 & \theta_i & w_k^\top & \theta_i w_k^\top \\
\frac{1}{\theta_i} w_k & w_k & \frac{1}{\theta_i} w_k w_k^\top & w_k w_k^\top \\
w_k & \theta_i w_k & w_k w_k^\top & \theta_i w_k w_k^\top
\end{bmatrix}.
\end{equation}
\end{small}
This transformation normalizes out the higher-order terms \( \theta_i^2 \), yielding linearized representations of all relevant bilinear expressions:

\begin{small}
\begin{equation}
\theta_i x_k^\top Q_x x_k = \mathrm{trace}(Q \cdot \theta_i w_k w_k^\top), \theta_i u_k^\top R_u u_k = \mathrm{trace}(R \cdot \theta_i w_k w_k^\top).
\end{equation}
\end{small}
The sparsity of each stage-wise matrix \( \tilde{Y}_{k,i} \) allows us to build a sparse block-diagonal relaxation for the full horizon problem. 

In the following, we introduce two categories of continuity constraints that arise from the multiple shooting discretization and the segment-wise time-scaling framework.
First, following the idea of direct multiple shooting, and in line with \eqref{direct shooting dynamics}, we enforce intra-segment continuity constraints to ensure consistency of the state trajectory across consecutive shooting intervals within the same segment. 
This is achieved by imposing the following equality constraints at the shooting nodes:
\begin{subequations} \label{intra-segment}
\begin{align}
w_k[i_a] - w_{k+1}[i_b] &= 0,  \\
\theta_i w_k[i_a] - \theta_i w_{k+1}[i_b] &= 0, \\
w_k/\theta_i[i_a] - w_{k+1}/\theta_i[i_b] &= 0,  \\
w_kw_k^\top[i_a,i_a] - w_{k+1}w_{k+1}^\top[i_b,i_b] &= 0, \label{} \\
\theta_i w_kw_k^\top[i_a,i_a] - \theta_i w_{k+1}w_{k+1}^\top[i_b,i_b] &= 0, \\
w_kw_k^\top/\theta_i[i_a,i_a] - w_{k+1}w_{k+1}^\top/\theta_i[i_b,i_b] &= 0, 
\end{align}
\end{subequations}
for $i =1, \ldots, M+1$ and where \( w_k \) and \( w_{k+1} \) denote the lifted variable vectors at consecutive shooting steps \( k \) and \( k+1 \), respectively. The index sets are defined using Python-style slicing: \( i_a = [n_x : 2n_x] \) refers to the predicted terminal state \( x_{k+1} \) from the interval \( [t_k, t_{k+1}] \), while \( i_b = [:n_x] \) refers to the initial state \( x_{k+1} \) of the next interval. These constraints ensure that the predicted and initialized states at each node are consistent across the entire segment. 

Second, we impose inter-segment continuity constraints to ensure that the terminal state of one segment aligns with the initial state of the next segment, like \eqref{segment continuity condition}. This is critical to ensure that the overall trajectory across all \( M+1 \) segments is both continuous and dynamically feasible. 
Specifically, we impose the equality constraints:
\begin{subequations} \label{inter-segment}
\begin{align}
w_{N_i-1,i}[i_a] - w_{0,i+1}[i_b] = 0, \label{} \\
w_{N_i-1,i}w_{N_i-1,i}^\top[i_a,i_a] - w_{0,i+1}w_{0,i+1}^\top[i_b,i_b] = 0, \label{}
\end{align}
\end{subequations}
for $i =1, \ldots, M$.
This lifting-based formulation preserves convexity while ensuring that the relaxed problem still encodes the structural consistency of the original \ac{OCP} trajectory.

For clarity of presentation and to enable a structured relaxation, we first introduce the following abstract representation of the original \ac{OCP} \eqref{original QCQP}:

\begin{small}
\begin{mini!}[1]
{w, \theta}{\sum_{i=0}^{M+1} w_t N_i\theta_i + \sum_{i=0}^{M+1}\sum_{k=0}^{N_i-1}\mathrm{trace}\left( \left(Q + R \right) \cdot \theta_i w_k w_k^\top \right)}
{\label{}}{}
\addConstraint{ w_{N_i-1,i}[i_a] - w_{0,i+1}[i_b]} {= 0, \quad i = 1, \dots, M}
\addConstraint{ Fw_k - K\theta_iw_k} {= 0, \quad i = 1, \dots, M+1}
\addConstraint{ Hw_k - h} {= 0, \quad i = 1, \dots, M+1} \label{}
\addConstraint{Gw_k - g} {\geq 0, \quad i = 1, \dots, M+1} \label{}
\addConstraint{ C\theta_i - c \geq 0, \theta_i } {\geq 0, \quad i = 1, \dots, M+1} \label{}
\addConstraint{ } { \quad  \quad  \quad  \quad   \quad   \quad   \quad   k = 0, \dots, N_i-1.  \notag} \label{}
\end{mini!}
\end{small}

Then, incorporating the lifted variable representation, block-wise structural constraints introduced above, we arrive at the following tailored and fast \ac{SDP} convex relaxation (Fast \ac{SDR}) tailored to the \ac{OCP} with characteristic time constraints:
\begin{algbox}{Fast semidefinite programming convex relaxation NLP}
\begin{small}
\begin{mini!}[1]
{\tilde{Y}}{\sum_{i=0}^{M+1} w_t N_i\theta_i + \sum_{i=0}^{M+1}\sum_{k=0}^{N_i-1}\mathrm{trace}\left( \left(Q + R \right) \cdot \theta_i w_k w_k^\top \right)}
{\label{relaxed MPCCT}}{}
\addConstraint{ \text{inter-segment constraints} \, } { \eqref{inter-segment}, \, \quad i = 1, \ldots, M \notag}
\addConstraint{ \text{intra-segment constraints} \,} { \eqref{intra-segment}, \, \quad i = 1, \ldots, M+1 \notag}
\addConstraint{ Fw_k - K\theta_i w_k} {= 0,  \quad i = 1, \ldots, M+1}
\addConstraint{ Fw_k/\theta_i  - Kw_k} {= 0,  \quad i = 1, \ldots, M+1}
\addConstraint{ Fw_kw_k^\top - K\theta_i w_kw_k^\top} {= 0,  \quad i = 1, \ldots, M+1}
\addConstraint{ Fw_kw_k^\top/\theta_i - Kw_kw_k^\top} {= 0,  \quad i = 1, \ldots, M+1}
\addConstraint{ Hw_k - h} {= 0,  \quad i = 1, \ldots, M+1} \label{}
\addConstraint{ Hw_k\theta_i - h\theta_i} {= 0,  \quad i = 1, \ldots, M+1} \label{}
\addConstraint{ Hw_k/\theta_i - h/\theta_i} {= 0,  \quad i = 1, \ldots, M+1} \label{}
\addConstraint{ Hw_kw_k^\top - hw_k^\top} {= 0,  \quad i = 1, \ldots, M+1} \label{}
\addConstraint{ H\theta_i w_kw_k^\top - h\theta_i w_k^\top} {= 0,  \quad i = 1, \ldots, M+1} \label{}
\addConstraint{ Hw_kw_k^\top/\theta_i - hw_k^\top/\theta_i} {= 0,  \quad i = 1, \ldots, M+1} \label{}
\addConstraint{Gw_k - g} {\geq 0,  \quad i = 1, \ldots, M+1} \label{}
\addConstraint{\theta_i (Gw_k - g)} {\geq 0,  \quad i = 1, \ldots, M+1} \label{}
\addConstraint{(Gw_k - g)/\theta_i} {\geq 0,  \quad i = 1, \ldots, M+1} \label{}
\addConstraint{C\theta_i - c} {\geq 0,  \quad i = 1, \ldots, M+1} \label{}
\addConstraint{\theta_i \geq 0, 1/\theta_i} {\geq 0,  \quad i = 1, \ldots, M+1} \label{}
\addConstraint{} { \quad  \quad   \quad   \quad   \quad   \quad  k = 0, \dots, N_i-1,\notag} \label{}
\end{mini!}
\end{small}
\end{algbox}

\noindent where the products $w_k/\theta_i, \theta_i w_k, w_k w_k ^\top, \theta_i w_kw_k^\top, w_kw_k^\top/\theta_i$ are replaced by the corresponding entries from the PSD matrix $\tilde{Y}$.
Note that in the lifted semidefinite formulation, each decision variable vector \( w_k \) is embedded into a corresponding positive semidefinite matrix \( \tilde{Y} \). 
As a result, any linear equality imposed on components of \( w_k \) translates into an affine constraint on specific subblocks of \( \tilde{Y} \).

In the proposed semidefinite relaxation, we incorporate additional constraints derived by multiplying existing linear constraints from the original \ac{NLP} formulation. While these constraints are theoretically redundant in the original nonconvex formulation, they serve a critical role in strengthening the convex relaxation by improving relaxation tightness and reducing the feasible set of the lifted problem~\cite{graesdal2024convexrelaxations}.
Thanks to the sparse and structure-preserving nature of the proposed \emph{Fast SDR} formulation, the obtained solution can be directly used as a high-quality initial guess for solving the original bilinear nonconvex problem. 
In particular, we demonstrate that the solution can be efficiently refined using off-the-shelf nonlinear programming solvers to recover dynamically consistent and near-optimal trajectories.

\section{Experimental Results}\label{} 
To comprehensively evaluate the proposed approach, this section presents experimental results across both simulation and real-world scenarios. 
We conduct a detailed performance comparison on a class of commonly studied benchmark problems. 
This enables a quantitative analysis of the solution, relaxation tightness, and computational efficiency in comparison to representative baselines. 
Subsequently, we deploy our method in a more complex real-world quadrotor waypoint flight with time-window constraints. 
Unlike conventional waypoint flight \cite{foehn2021time,hanover2024autonomous} where the quadrotor may pass waypoints at any time, our formulation imposes strict time window constraints at each waypoint, adding a significant layer of complexity.

\subsection{Numerical Comparison and Analysis}

In this section, we evaluate the proposed SDP-based convex relaxation approach by applying it to a widely studied benchmark in eco-driving problems: the point-mass system~\cite{faris2025optimization, sabouni2024optimal}. 
The dynamics and cost structure of this system follow the canonical form of second-order linear motion, and are well-suited to test \ac{OCP} formulations under spatio-temporal constraints.
The system constraints and corresponding parameters used in our experiments are defined as follows. The state vector \( x(t) = [s(t), v(t)]^\top \in \mathbb{R}^2 \) consists of the position \( s(t) \) and velocity \( v(t) \), while the control input is the acceleration \( u(t) \in \mathbb{R} \). The dynamics of the system are governed by the standard double integrator model $\dot{s}(t) = v(t), \quad \dot{v}(t) = u(t)$.
The objective is to minimize the following cost $J = t_f + \frac{1}{2}\int_0^{t_f} u^2(t) \mathrm{d}t $.
The system is subject to box constraints on both velocity and acceleration $v_{\min} \leq v(t) \leq v_{\max}, \quad u_{\min} \leq u(t) \leq u_{\max}$ with \( v_{\min} = 0 \)~m/s, \( v_{\max} = 2 \)~m/s, \( u_{\min} = -1 \)~m/s\(^2\), and \( u_{\max} = 1 \)~m/s\(^2\). 
The terminal time \( t_f \) is treated as a free optimization variable under the time-scaling formulation presented earlier.
Characteristic time constraints are incorporated into the problem. 
In particular, one characteristic time \( \tau_1 \) is defined to correspond to a specific spatial location along the trajectory. 
In our setup, the characteristic time constraint is activated at the point where the travel distance reaches \( s(\tau_1) = 0.6 \).
The terminal conditions for the system are fixed as \(x(t_f) = [1, 0]^\top\).

We use the commercial SDP solver \texttt{MOSEK} \cite{mosek} to solve the problem \eqref{relaxed MPCCT}.
To comprehensively evaluate the effectiveness of our proposed approach, we focus on three aspects of performance: \textit{relaxation gap}, \textit{optimality}, and \textit{computational efficiency}. 
All experiments were conducted on a laptop equipped with an Apple M4 processor and 16 GB of RAM.

\begin{figure}[ht]
    \centering
    \includegraphics[width=\linewidth]{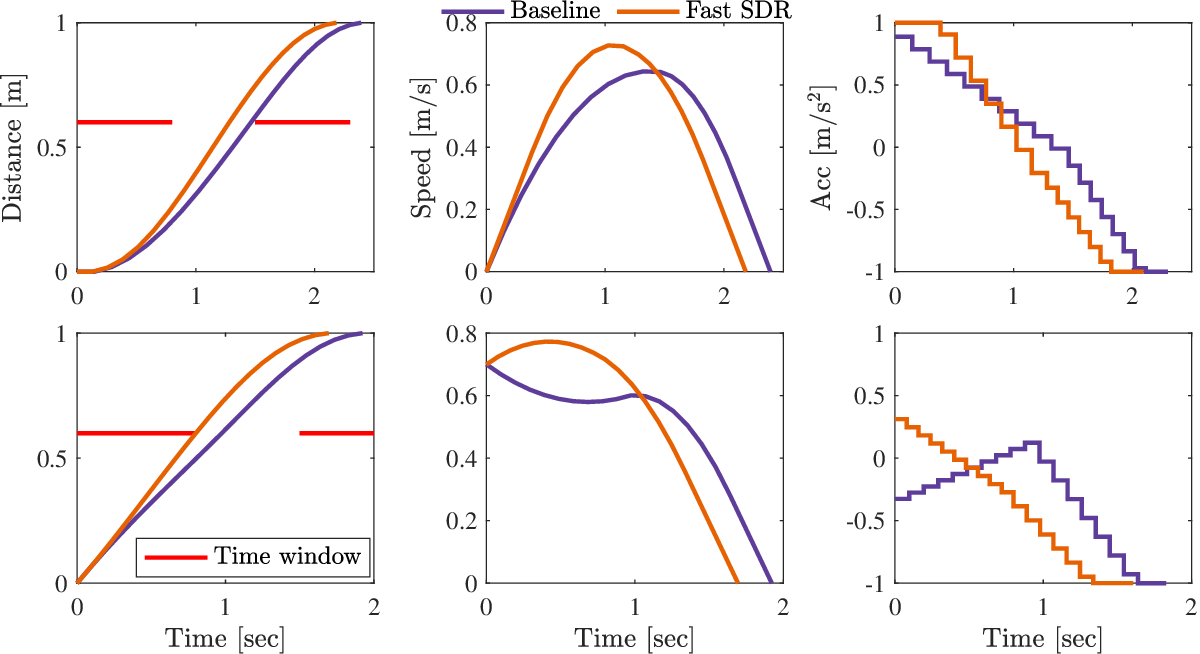}
    \caption{Representative trajectories with active upper (top) and lower (bottom) characteristic time constraints.}
    \label{fig:relaxation_cases}
\end{figure}

Firstly, to evaluate the tightness of the convex relaxation, we define the following relaxation gap metric $\sigma_{\text{rel}} = \frac{J_{\text{RCVX}} - J_{\text{CVX}}}{J_{\text{RCVX}}}$, where \( J_{\text{CVX}} \) is the objective value obtained by solving the convex SDR problem, and \( J_{\text{RCVX}} \) is the value obtained by refining the SDR solution using \texttt{IPOPT} \cite{wachter2006implementation} on the original \ac{NLP} problem. 
The refinement is initialized with the solution trajectory produced by the convex relaxation. 
This two-step procedure yields a high-quality feasible solution, allowing us to quantify how close the SDR relaxation is to global optimality.
Table~\ref{tab:results} reports statistical results across various initializations. 
In addition, Fig.~\ref{fig:relaxation_cases} visualizes two representative scenarios, one in which the characteristic time upper bound is active, and another one where the lower bound is active. 
The results demonstrate that for the double integrator problem, the proposed convex relaxation yields a tight approximation of the globally optimal solution refined by \texttt{IPOPT}. 
Notably, the relaxation gap remains consistently below 1\% across most cases, indicating that the SDR closely captures the structure of the original problem.

To further assess the quality of our proposed approach, we benchmark it against a state-of-the-art baseline: the explicit interior-point constraint formulation introduced in~\cite{ard2025energy}, which has been widely adopted in recent studies for handling characteristic time constraints \cite{han2019fundamentals}. 
This baseline enforces temporal constraints by explicitly anchoring the system's state to occur at prescribed time instants. 
However, due to its reliance on fixed characteristic time assignments, the method inherently imposes assumptions on timing feasibility, which often leads to suboptimal solutions.
Our comparative analysis highlights the improved optimality of the proposed semidefinite relaxation framework under the same conditions.
We define the optimality comparison metric $\delta_{\text{opt}} = \frac{J_{\text{BASE}}-J_{\text{RCVX}} }{J_{\text{RCVX}} }$, where \( J_{\text{BASE}} \) denotes the cost returned by the baseline methods. 
The results demonstrate a relative deviation between the proposed method and the baseline ranging from 2.5\% to 8.9\%.
From the results, we can see that solutions obtained via the proposed semidefinite relaxation and refinement pipeline consistently outperform the baseline approaches, especially in scenarios with tight and active time constraints. 
This confirms that the convex relaxation preserves global structure and enables recovery of superior feasible trajectories.

\begin{figure}[ht]
    \centering
    \includegraphics[width=\linewidth, trim=0cm 0.12cm 0cm 0.0cm, clip]{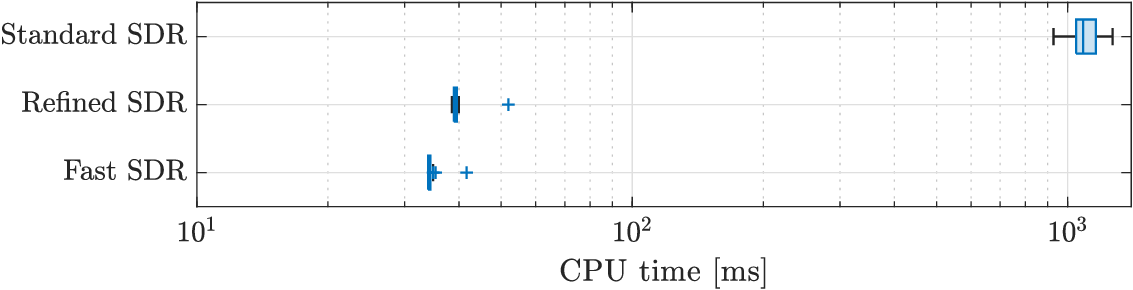}
    \caption{Comparison of computation time across \emph{Standard SDR}, \emph{Fast SDR}, and \emph{Refined SDR}.}
    \label{fig:runtime}
\end{figure}

We then compare the computational cost across three different methods. 
The first, termed \emph{Standard SDR}, refers to the conventional SDR formulation \cite{parrilo2003semidefinite,yang2025new} solved using the \texttt{MOSEK} solver, without exploiting any problem-specific sparsity. 
The second, \emph{Fast SDR}, corresponds to our tailored semidefinite program that leverages the inherent sparse structure of the \ac{OCP}, significantly reducing the number of lifting variables and computation time. 
The third approach, denoted as \emph{Refined SDR}, is a two-stage refinement pipeline in which Fast SDR is first used to generate a high-quality initial guess, which is then refined using \texttt{IPOPT} applied to the \ac{NLP} problem.
Fig.~\ref{fig:runtime} reports the average runtime of the three methods. 
The results demonstrate that the \emph{Fast SDR} achieves at least an order-of-magnitude speedup over the standard SDR formulation, owing to its reduced matrix dimensions and problem-specific structure.
The total runtime of \emph{Refined SDR} includes both the time to solve the convex relaxation problem, i.e., the \emph{Fast SDR}, and the time to solve the original \ac{NLP} using \texttt{IPOPT} warm-started with the \emph{Fast SDR} solution.
Although \emph{Refined SDR} requires a two-step solve (\emph{Fast SDR} + \texttt{IPOPT}), the total runtime remains competitive, as the \ac{NLP} solver benefits from a high-quality initialization.

\begin{table}[htbp]
\centering
\caption{Statistical Comparison of Performance Metrics Across Varying Initial Conditions}
\label{tab:results}
\scalebox{0.8}{ 
\begin{tabular}{l *{7}{S[table-format=1.4]}}
\toprule
\textbf{} & \multicolumn{6}{c}{ Different initial speed $v(0)$ values} \\
\cmidrule(l){2-8}
 & {$0$} & {$0.2$} & {$0.3$}  & {$0.5$} & {$0.7$} & {$0.9$} & {$1.0$}\\
\midrule
$J_{\text{CVX}}$    & 2.789  & 2.492  & 2.366    & 2.159   & 2.017  & 2.024 & 2.052 \\
$J_{\text{RCVX}}$   & 2.789  & 2.492  & 2.366    & 2.159   & 2.017  & 2.042 & 2.131 \\
$J_{\text{BASE}}$   & 2.859  & 2.63   & 2.504    & 2.352   & 2.185  & 2.187 & 2.024 \\
\midrule
\rowcolor{gray!10} 
$\sigma_{\text{rel}}$ (\%)     & 0 & 0 & 0  & 0  & 0 & 0.88 & 3.70\\
\rowcolor{gray!25} 
$\delta_{\text{opt}}$ (\%) & 2.51 & 5.54 & 5.83 & 8.94 & 8.33 & 7.10 & 7.95 \\
\bottomrule
\end{tabular}
}
\end{table}

\subsection{Quadrotor Waypoint Flight Experimental Test}

\begin{table}[t]
\centering
\caption{Waypoints and passing times for the two scenarios.}
\renewcommand{\arraystretch}{1.2}
\setlength{\tabcolsep}{6pt}
\resizebox{\linewidth}{!}{
\begin{tabular}{c c c c c}
\hline
 & \multicolumn{2}{c}{Scenario 1}  & \multicolumn{2}{c}{Scenario 2} \\ 
 & Position & Time Win. & Position & Time Win. \\ \hline
Waypoint 1 & (0.16, -1.03, 1.12) & [0, 0.5] & (0.17, -1.17, 1.33) & [0, 0.5] \\
Waypoint 2 & (2.36, 0.55, 1.69) & [1.8, 3] & (2.42, 0.60, 1.35) & [1.8, 3] \\
Waypoint 3 & (0.87, 1.80, 1.27) & [2.5, 3] & (0.78, 1.76, 1.89) & [2.5, 3] \\
Waypoint 4 & (-1.87, 0.28, 1.63) & [4, 4.5] & (-1.89, 0.30, 1.26) & [3.5, 4] \\ \hline
Pass Times & \multicolumn{2}{c}{(0.49, 1.96, 2.58, 4.00)} & \multicolumn{2}{c}{(0.49, 1.94, 2.51, 3.50)} \\ \hline
\end{tabular}}
\label{tab:waypoints}
\end{table}

\begin{figure}[t]
    \centering
    \includegraphics[width=\linewidth]{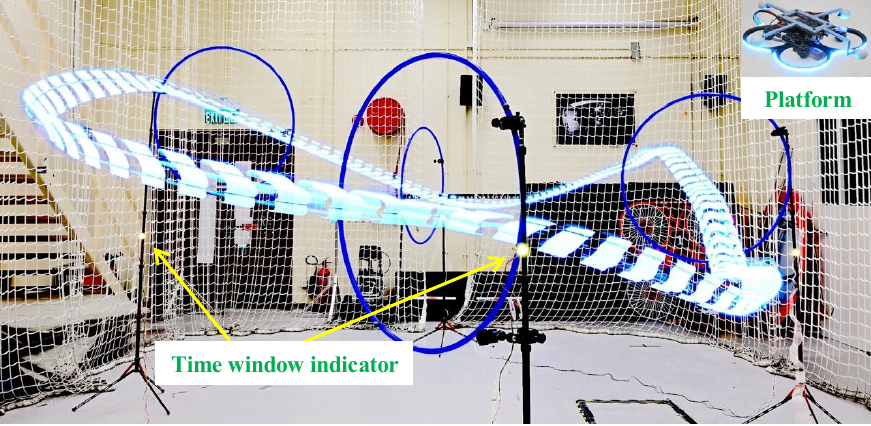}
    \caption{Time-lapse of an experiment showing the trajectory executed by the quadrotor. To better visualize the evolution of the time windows during the experiments, an indicator is placed beneath each waypoint: a green light \textcolor{green}{$\boldsymbol{\bullet}$} signified that the waypoint was within its open time window and could be safely crossed, whereas a red light \textcolor{red}{$\boldsymbol{\bullet}$} indicated that passage through the waypoint was temporarily prohibited. }
    \label{fig:Platform}
\end{figure}

To further validate the effectiveness of the proposed method, we conducted a set of real-world experiments on a more challenging quadrotor waypoint flight under time window constraints. 
The experiments were carried out using a quadrotor platform operating within a Vicon motion capture environment, where the quadrotor was required to pass 4  waypoints in three-dimensional space while satisfying waypoint-specific open time windows.
As illustrated in Fig. \ref{fig:Platform}, the experimental setup consists of an Intel NUC 12 with Core i5-1250P computing unit with Wi-Fi module for calculating control commands using low-level model predictive control and transmitting them to the PX4 firmware\footnote{\url{https://px4.io/}}. 
The proposed fast semidefinite relaxation-based trajectory planner was implemented in the planning layer to generate optimal waypoint flight trajectories that respect both spatial and temporal constraints.

\begin{figure}[t]
    \centering
    \includegraphics[width=\linewidth]{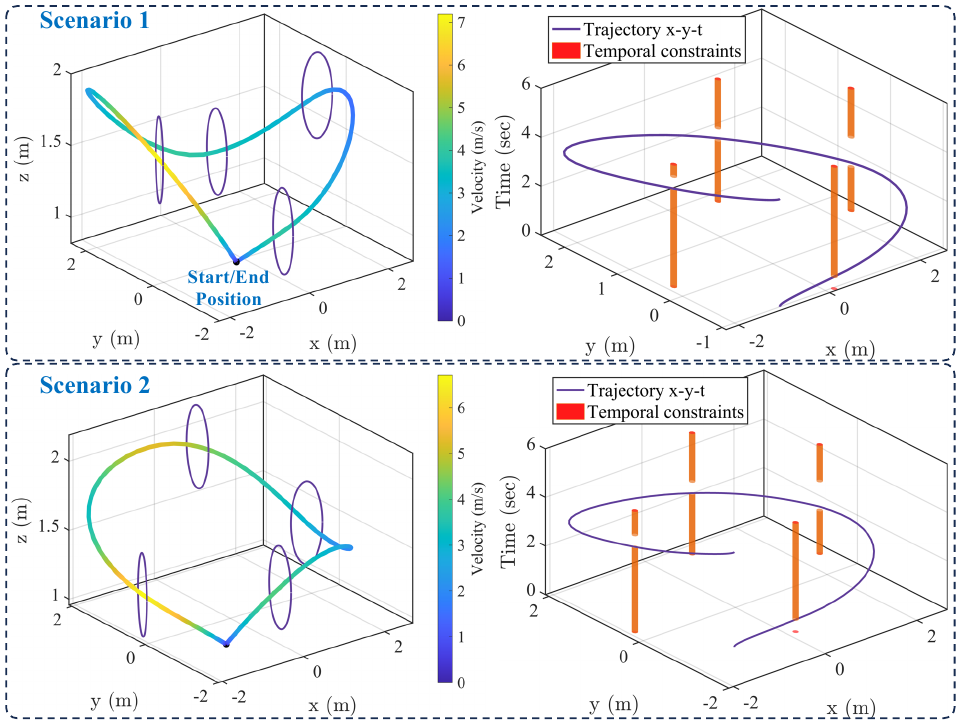}
    \caption{Trajectories obtained from real-world experiments in two scenarios. On the left, the position with the velocity color scheme is shown. On the right, the projected horizontal position is plotted over time on the z-axis. The orange bars visualize the open or closed temporal constraints.}
    \label{fig:traj_time}
\end{figure}

\begin{figure}[t]
    \centering
    \includegraphics[width=\linewidth]{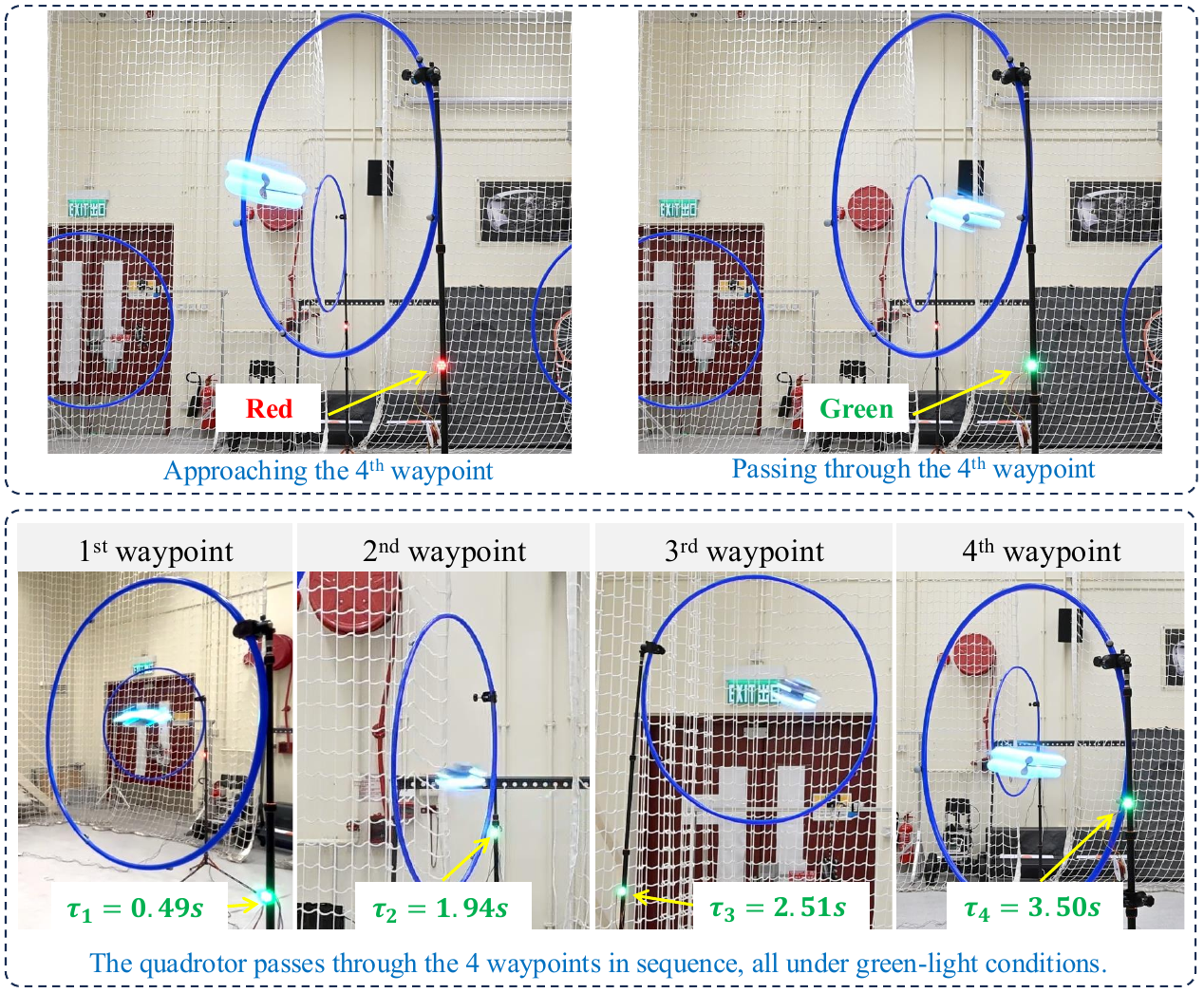}
    \caption{The snapshots of the actual flight trajectories (top: scenario 1, bottom: scenario 2). Video is available at https://youtu.be/ZDcIhLDWEqA.}
    \label{fig:test}
\end{figure}

In comparison, two distinct flight scenarios were conducted in real-world environments to evaluate the capability and robustness of the proposed system. 
The 3D positions of the waypoints were randomly selected, and different temporal windows were assigned to each waypoint to impose spatio-temporal constraints. 
The specific waypoint locations and their corresponding time window settings for the two flight scenarios are summarized in Tab.~\ref{tab:waypoints}.
The results of the two flight experiments are illustrated in Fig. \ref{fig:traj_time} and Fig. \ref{fig:test}. 
The figures depict the quadrotor trajectories, color-coded by instantaneous speed, and capture variations in the time windows as the quadrotor approaches and passes through each waypoint from different perspectives. 
To provide a more intuitive understanding of the proposed method’s effectiveness, multiple viewpoints were selected to highlight the dynamic interaction between trajectory evolution and the active time-window constraints.
In Scenario 1, we focus on the passage through the 4th waypoint as a representative case. 
Several sequential frames captured around this event clearly show the rapid trajectory transition during the approaching and crossing phases. 
As visualized in the top of Fig. \ref{fig:test}, the corresponding indicator below the waypoint switches from red to green precisely when the quadrotor arrives, signifying the transition from the closed to the open time window. 
This synchronization demonstrates that the quadrotor crossed the waypoint exactly at the feasible instant, effectively minimizing flight time while maintaining constraint satisfaction. 
The recorded waypoint crossing times in Tab.~\ref{tab:waypoints} further confirm that the quadrotor passed each waypoint safely and precisely within its respective open interval.
In Scenario 2, the full flight through all 4 waypoints is illustrated. 
The resulting trajectories strictly comply with all  spatio-temporal constraints throughout the entire flight. 
The planned trajectory dynamically adapts to the activation of each time window, maintaining feasibility even under tighter temporal restrictions. 
The waypoint crossing times listed in Table~2 again validate that the quadrotor consistently achieved on-time passage across all waypoints without violating any time-window constraint. 
These results collectively demonstrate the real-world feasibility of the proposed approach.

\acresetall 
\section{Conclusion}
This paper presented a novel convex relaxation framework for solving \acp{OCP} with spatio-temporal constraints, where both waypoint locations and time-window constraints must be simultaneously satisfied. 
We first introduced a compact time-scaling direct multiple shooting scheme to transform it into a structured yet (nonconvex) \ac{NLP} problem. 
To efficiently handle the inherent nonconvexity, a fast \ac{SDP}-based convex relaxation was developed, which exploits the sparse coupling structure between state, control, and timing variables. 
Comprehensive numerical evaluations demonstrate that the proposed approach yields a tight approximation of the original problem, achieves up to an 8.9\% improvement in optimality, and provides an order-of-magnitude reduction in computational time.
Furthermore, real-world experiments on a quadrotor waypoint flight task under time-window constraints validated the method’s practical feasibility. 
The quadrotor successfully executed on-time waypoint crossings, confirming that the proposed convex relaxation framework can be effectively deployed on physical platforms.
Future work will incorporate more complex quadrotor dynamics to fully leverage the actuators’ capabilities and enable on-board applications on resource-limited aerial and ground robots.


\bibliographystyle{IEEEtran}

\bibliography{cas-refs.bib}   

\end{document}